\documentclass{article}

\usepackage{arxiv}
\usepackage[utf8]{inputenc} 
\usepackage[T1]{fontenc}    
\usepackage{hyperref}       
\usepackage{url}            
\usepackage{booktabs}       
\usepackage{amsfonts}       
\usepackage{nicefrac}       
\usepackage{microtype}      
\usepackage{lipsum}
\usepackage{graphicx}
\usepackage{amsmath}
\usepackage{booktabs}

\graphicspath{ {./images/} }

\title{LGAI-Embedding-Preview Technical Report}

\author{LG AI Research\thanks{The complete list of authors who contributed to this work can be found in Appendix A.}}

\begin{document}
\maketitle

\begin{abstract}
This report presents a unified instruction-based framework for learning generalized text embeddings optimized for both information retrieval (IR) and non-IR tasks. Built upon a decoder-only large language model (Mistral-7B), our approach combines in-context learning, soft supervision, and adaptive hard-negative mining to generate context-aware embeddings without task-specific fine-tuning. Structured instructions and few-shot examples are used to guide the model across diverse tasks, enabling strong performance on classification, semantic similarity, clustering, and reranking benchmarks.

To improve semantic discrimination, we employ a soft labeling framework where continuous relevance scores—distilled from a high-performance dense retriever and reranker—serve as fine-grained supervision signals. In addition, we introduce adaptive margin-based hard-negative mining, which filters out semantically ambiguous negatives based on their similarity to positive examples, thereby enhancing training stability and retrieval robustness.

Our model is evaluated on the newly introduced MTEB (English, v2) benchmark, covering 41 tasks across seven categories. Results show that our method achieves strong generalization and ranks among the top-performing models by Borda score, outperforming several larger or fully fine-tuned baselines. These findings highlight the effectiveness of combining in-context prompting, soft supervision, and adaptive sampling for scalable, high-quality embedding generation.
\end{abstract}

\section{Introduction}
Text embeddings serve as the foundational representation for a wide range of natural language processing (NLP) tasks, including retrieval, classification, reranking, and semantic similarity. With the rise of large language models (LLMs), recent research has increasingly focused on adapting LLMs as general-purpose text encoders. Existing approaches such as Llama2Vec~\cite{behnamghader2024llm2vec}, E5-Mistral~\cite{wang2023improving}, and NV-Embed~\cite{lee2024nv} demonstrated significant improvements in performance by fine-tuning decoder-only LLMs with synthetic data, architectural modifications, and carefully designed pooling strategies. However, these models often rely on full model fine-tuning or structural changes, which can limit scalability and overlook the inherent generalization capacity of LLMs.

In response to these limitations, there has been a growing interest in using in-context learning (ICL) to generate embeddings without modifying model weights. Recent methods such as BGE-en-icl~\cite{li2024making} show that structured prompts and few-shot examples can effectively guide LLMs to produce high-quality embeddings across diverse tasks. This paradigm shift emphasizes flexibility and resource efficiency, enabling models to adapt to multiple use cases with minimal tuning.

Another challenge concerns the reliability of labels and signals used to supervise the model during training. Traditional contrastive learning approaches often suffer from noisy or insufficient labels, particularly due to the presence of false negatives. To address this, we incorporate soft labeling strategies by distilling continuous-valued relevance scores from a high-performance retrieval pipeline. These soft targets offer nuanced semantic signals that guide the model toward better alignment with human similarity judgments. In addition, we adopt adaptive hard-negative mining techniques that dynamically select informative negatives based on their similarity to the corresponding positives, reducing the impact of semantically overlapping examples.

In this work, we present LGAI-Embedding-Preview, a generalized embedding model built upon Mistral-7B and enhanced with in-context learning, soft supervision, and adaptive hard-negative sampling. Without modifying the underlying model architecture, our approach achieves strong generalization across a wide range of retrieval and non-retrieval tasks. Evaluated on the newly introduced MTEB(English, v2) benchmark~\cite{enevoldsen2025mmteb}, our model demonstrates competitive performance across 41 diverse tasks, ranking among the top systems in terms of Borda score. Our findings highlight the effectiveness of combining in-context LLM capabilities with high-quality training signals for scalable and accurate embedding generation.

\section{Related Works}
\label{sec:headings}
\paragraph{LLM-based Text Embedding Models}
Recent research has increasingly explored the use of large language models (LLMs) as backbone encoders for text embedding tasks. This shift is evident in models such as Llama2Vec\cite{behnamghader2024llm2vec}, which introduced two pretraining objectives to better align LLMs with embedding tasks, yielding substantial performance improvements on retrieval benchmarks like BEIR. However, its performance on the MTEB leaderboard remains relatively modest. Other models such as E5-Mistral\cite{wang2023improving}, Linq\cite{LinqAIResearch2024}, and Gecko\cite{lee2024gecko} have leveraged large-scale synthetic data to effectively fine-tune LLMs, achieving strong results across both retrieval and non-retrieval tasks. NV-Embed\cite{lee2024nv} further advances this line of work by incorporating a latent attention pooling mechanism and a two-stage training strategy to mitigate false negatives, leading to significant improvements in retrieval robustness.

\paragraph{Shift Toward In-Context Learning}
Despite the strong performance of LLM-based embedding models, prior approaches have often relied heavily on architectural modifications—such as replacing unidirectional attention with bidirectional mechanisms—or full model fine-tuning. These strategies, while effective, tend to overlook the inherent generalization capabilities of LLMs and require substantial computational resources. Recently, however, there has been a shift toward leveraging in-context learning (ICL) as a more efficient alternative. Models such as BGE-en-icl~\cite{li2024making} demonstrate that task-specific prompts and demonstrations can be used to condition LLMs for embedding generation without modifying model weights. This emerging paradigm highlights the potential of ICL for building flexible and adaptive embedding systems that generalize well across tasks while minimizing the cost of training and deployment.

\paragraph{High-Quality Supervision via Soft Labeling and Hard-Negative Mining.}
Recent advances in embedding model training highlight the importance of soft labeling and hard-negative mining. In particular, soft supervision signals derived from high-capacity reranker models—often implemented via teacher-student frameworks—have proven effective in guiding embedding models toward better alignment with semantic similarity objectives~\cite{mandal2024theoretical}. Meanwhile, hard-negative mining strategies play a critical role in closing the semantic gap between positive and negative pairs and in mitigating the risk of false negatives during contrastive learning. Building on these insights, our goal is to develop a generalized embedding model that preserves the inherent strengths of LLMs by leveraging in-context learning. Without requiring architectural modifications or full-scale fine-tuning, our approach aims to achieve strong adaptability across both retrieval and non-retrieval tasks, while maintaining efficiency in training and deployment.

\begin{table*}[t]
\centering
\caption{Instruction templates for the training datasets used in our experiments}
\resizebox{\textwidth}{!}{%
\begin{tabular}{p{5.5cm} p{11cm}} 
\hline
\textbf{Task Name} & \textbf{Instructions} \\
\hline
ArguAna & Given a claim, find documents that refute the claim.  \\
ELI5 & Provided a user question, retrieve the highest voted answers on Reddit ELI5 forum. \\
FEVER & Given a claim, retrieve documents that support or refute the claim. \\
FiQA2018 & Given a financial question, retrieve user replies that best answer the question. \\
HotpotQA & Given a multi-hop question, retrieve documents that can help answer the question. \\
MSMARCO & Given a web search query, retrieve relevant passages that answer the query.  \\
Natural Question & Given a question, retrieve Wikipedia passages that answer the question. \\
QuoraDupQuestion & Given a question, retrieve questions that are semantically equivalent to the given question. \\
SQuAD & Given a question, retrieve passages that answer the question \\
STS12, STS22, STSBenchmark & Retrieve semantically similar text. \\
AmazonCounterfactualClassification & Classify a given Amazon customer review text as either counterfactual or not-counterfactual. \\
AmazonReviewsClassification & Classify the given Amazon review into its appropriate rating category. \\
Banking77Classification & Given a online banking query, find the corresponding intents.  \\
EmotionClassification & Classify the emotion expressed in the given Twitter message into one of the six emotions: anger, fear, joy, love, sadness, and surprise. \\
ImdbClassification & Classify the sentiment expressed in the given movie review text from the IMDB dataset. \\
MTOPIntentClassification & Classify the intent of the given utterance in task-oriented conversation. \\
ToxicConversationsClassification & Classify the given comments as either toxic or not toxic.  \\
TweetSentimentExtractionClassification & Classify the sentiment of a given tweet as either positive, negative, or neutral. \\
ArxivClusteringP2P & Identify the main and secondary category of Arxiv papers based on the titles and abstracts. \\
ArxivClusteringS2S & Identify the main and secondary category of Arxiv papers based on the titles. \\
BiorxivClusteringP2P & Identify the main category of Biorxiv papers based on the titles and abstracts.  \\
BiorxivClusteringS2S & Identify the main category of Biorxiv papers based on the titles. \\
MedrxivClusteringP2P & Identify the main category of Medrxiv papers based on the titles and abstracts. \\
MedrxivClusteringS2S & Identify the main category of Medrxiv papers based on the titles. \\
RedditClustering &  Identify the topic or theme of Reddit posts based on the titles. \\
RedditClusteringS2S & Identify the topic or theme of Reddit posts based on the titles and posts. \\
StackexchangeClustering &  Identify the topic or theme of StackExchange posts based on the titles. \\
StackexchangeClusteringP2P & Identify the topic or theme of StackExchange posts based on the given paragraphs. \\
TwentyNewsgroupsClustering & Identify the topic or theme of the given news articles. \\
SciDocsRR & Given a title of a scientific paper, retrieve the titles of other relevant papers. \\
StackOverflowDupQuestions & Retrieve duplicate questions from StackOverflow forum. \\
\hline
\end{tabular}
}
\label{tab:training_samples}
\end{table*}

\section{Training Dataset}
\label{sec:trainingdataset}
\paragraph{Retrieval Datasets} Following the common approaches taken by top-performing models on the English MTEB leaderboard \cite{enevoldsen2025mmteb}, we employ a set of publicly available retrieval datasets, including MSMARCO \cite{nguyen2016ms}, HotpotQA \cite{yang2018hotpotqa}, Natural Questions \cite{kwiatkowski2019natural}, SQuAD \cite{rajpurkar2016squad}, ELI5 \cite{fan2019eli5}, ArguAna \cite{wachsmuth2018retrieval}, FiQA \cite{maia201818}, FEVER \cite{thorne2018fever}, Quora Duplicate Questions \cite{sharma2019natural}. These retrieval datasets are annotated not only with hard labels, such as positive and negative pairs, but are also enhanced with soft labels, which are described in detail in the following section \ref{sec:method1}. Furthermore, they are processed through a sophisticated hard-negative mining strategy, which will also be detailed in the subsequent section \ref{sec:method3}.

\paragraph{Non-Retrieval Datasets} Similar to other models \cite{lee2024nv,li2024making}, we also incorporate publicly available datasets from non-retrieval benchmarks, particularly those associated with classification, clustering, reranking, and semantic textual similarity (STS) within the MTEB tasks. Importantly, we use only the training data and omit test splits to maintain evaluation integrity.

For classification tasks, we utilize a range of benchmark datasets, including AmazonCounterfactualClassification \cite{o2021wish}, AmazonReviewsClassification \cite{mcauley2013hidden}, Banking77Classification \cite{casanueva2020efficient}, EmotionClassification \cite{saravia2018carer}, ImdbClassification \cite{maas2011learning}, MTOPIntentClassification \cite{li2020mtop}, ToxicConversationsClassification \cite{jigsaw-unintended-bias-in-toxicity-classification}, and TweetSentimentExtractionClassification \cite{tweet-sentiment-extraction}. 

We utilize a range of clustering datasets such as ArxivClustering\footnote{https://www.kaggle.com/datasets/Cornell-University/arxiv}, BiorxivClustering \footnote{https://api.biorxiv.org/}, MedrxivClustering\footnote{https://api.medrxiv.org/}, TwentyNewsgroupsClustering \cite{lang1995newsweeder}, RedditClustering \cite{geigle:2021:arxiv}, StackExchangeClustering \cite{geigle:2021:arxiv}.

For reranking tasks, we incorporate SciDocsRR \cite{cohan2020specter} and StackOverflowDupQuestions \cite{liu2018linkso} into our training corpus. Additionally, we include the training splits of three widely used semantic textual similarity datasets—STS12 \cite{agirre2012semeval}, STS22 \cite{chen-etal-2022-semeval}, and STS-Benchmark \cite{chen-etal-2022-semeval}—to enhance the model’s ability to capture fine-grained semantic relations.

We summarize the training datasets and the instructions used for each in Table \ref{tab:training_samples}, which are adapted from prior work \cite{wang2023improving,li2024making}.

\paragraph{Data Conversion} Due to the limited availability of publicly released semantic textual similarity (STS) training datasets, we repurpose natural language inference (NLI) data as a surrogate for STS supervision in this study. 

\begin{figure}
  \centering
  \includegraphics[width=0.5\linewidth]{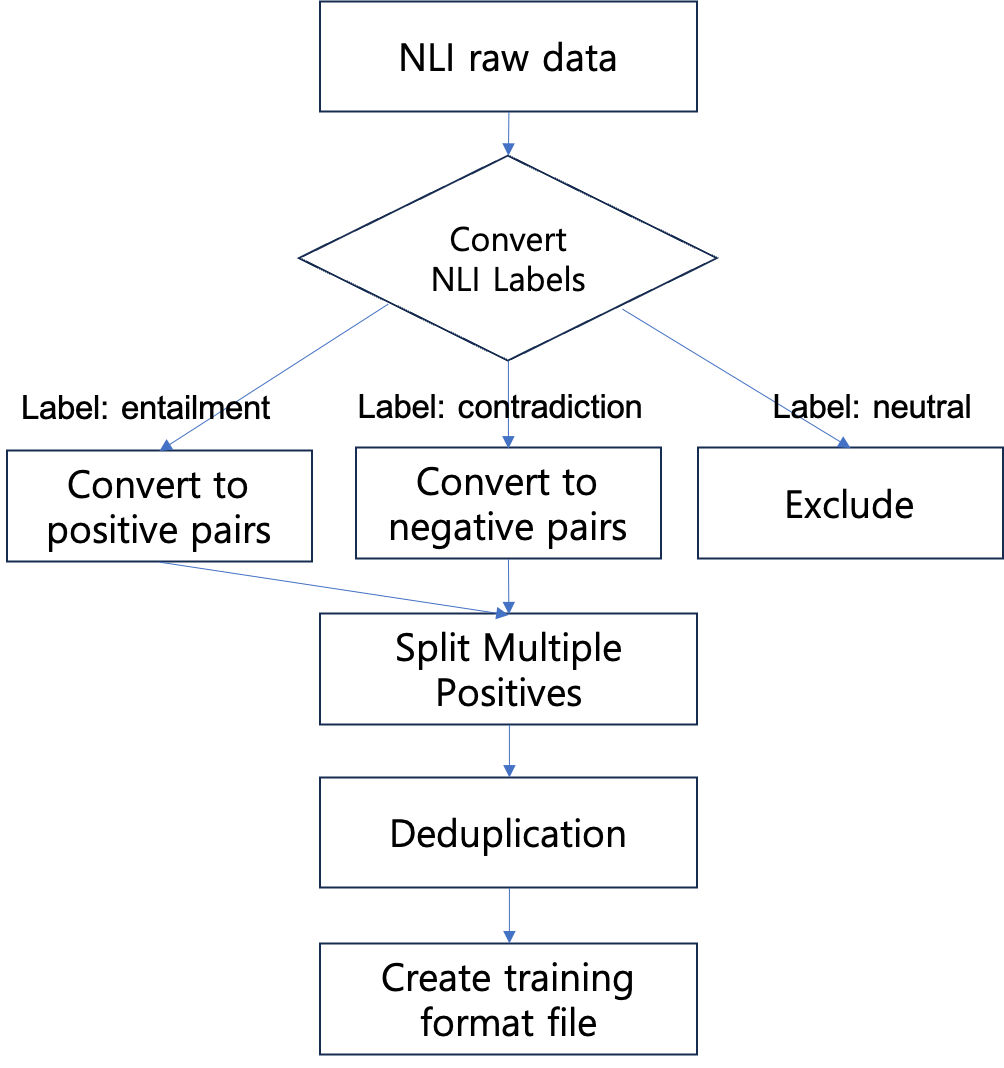}
  \caption{Flowchart illustrating the conversion process from Natural Language Inference (NLI) data to STS-style training examples. Sentence pairs labeled as “entailment” are retained and assigned high similarity scores, while “contradiction” pairs are converted into negative pairs with low similarity scores. Pairs labeled as “neutral” are discarded due to their ambiguous semantic alignment. The resulting examples are reformatted to match the STS input format and used as training data for similarity modeling.}
  \label{fig:datafusion}
\end{figure}

As illustrated in Figure~\ref{fig:datafusion}, we begin with a collection of NLI sentence pairs, each labeled as either “entailment,” “neutral,” or “contradiction.” Based on the original NLI label, each sentence pair is mapped to an approximate similarity score: entailment pairs are assigned high similarity, contradiction pairs are treated as negative examples with low similarity, and neutral pairs are discarded due to their semantic ambiguity.

To ensure consistency with the STS task format, neutral pairs are discarded due to their semantic ambiguity, while entailment and contradiction pairs are retained. The resulting sentence pairs are then reformatted to align with the structure of standard STS training data and incorporated into the training corpus for contrastive learning. Soft similarity scores are also added, as described in Section \ref{sec:method1}. 

To handle cases where a single query is associated with multiple positive passages, each query-positive pair is separated into individual training instances. For example, if a query has two positive passages, \texttt{[“A1”, “A2”]}, it is transformed into two distinct examples: \texttt{{query: “A”, positive: “A1”}} and \texttt{{query: “A”, positive: “A2”}}.
Following this, a deduplication step is applied to remove redundant \texttt{(query, positive)} pairs from the dataset. This process eliminates repeated entries and contributes to improved training efficiency by reducing unnecessary data redundancy.

\section{Proposed Methods}
\label{sec:method}
\subsection{Knowledge Distillation via Soft Labeling}
\label{sec:method1}
To enhance the semantic discrimination capability of our embedding model, we employ soft labeling supervision using similarity scores generated by our retrieval pipeline system. These continuous-valued relevance signals serve as soft targets, guiding the embedding model to learn nuanced semantic relationships through contrastive learning.

Unlike hard labels, which provide binary relevance judgments, soft labels offer richer information by capturing the teacher model's confidence across different candidates. This approach aligns with recent theoretical insights: Mandal et al. 
 \cite{mandal2024theoretical} demonstrate that soft label supervision allows student models to generalize more effectively and requires fewer neurons to approximate the teacher's decision boundaries compared to hard-labeled training targets. 

\begin{figure}
  \centering
  \includegraphics[width=0.8\linewidth]{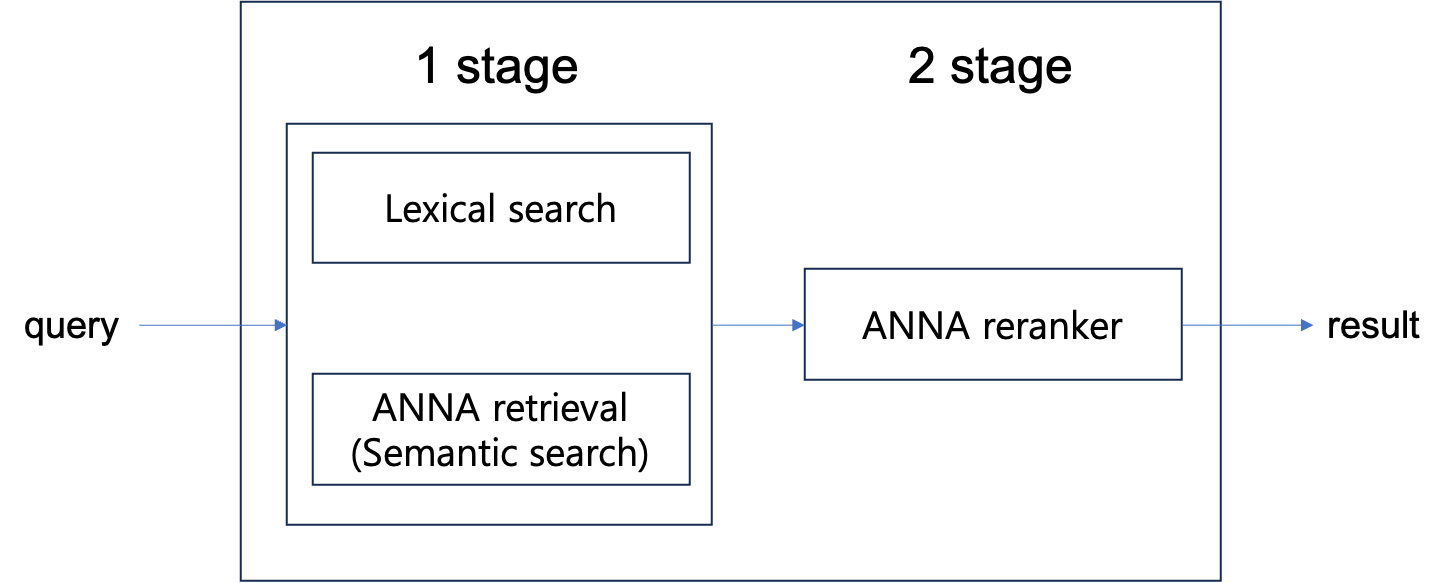}
  \caption{Overview of the ANNA IR pipeline. The system employs a two-stage retrieval framework. In the first stage, candidate passages are retrieved using both lexical search and semantic search (ANNA retrieval). In the second stage, a cross-encoder reranker refines the results by scoring the query-document pairs for final ranking.}
  \label{fig:fig1}
\end{figure}

For soft labeling in this study, we employ our in-house retrieval system, the ANNA IR pipeline, as illustrated in Figure~\ref{fig:fig1}. The pipeline integrates both lexical and semantic retrieval components, followed by a reranker to generate relevance scores. The ANNA IR pipeline is a two-stage retrieval system. In the first stage, candidate passages are retrieved using both lexical search and semantic retrieval (ANNA retrieval) \cite{jun2022anna}. This hybrid approach ensures coverage of both surface-level and semantic matches. In the second stage, the retrieved candidates are re-ranked using the ANNA reranker, which assigns fine-grained relevance scores based on contextual alignment with the query. The final ranked list is used for downstream tasks or soft label supervision.

We employ three scoring functions to estimate the relevance between a query $q$ and a document $d$:

\begin{itemize}
    \item \textbf{Lexical Search (BM25):} \\
    The BM25 score is computed based on exact token overlap, using the formula:
    \[
    \text{Lexical score}(d, q) = \sum_{i=1}^{n} \text{IDF}(q_i) \cdot \frac{f(q_i, d) \cdot (k_1 + 1)}{f(q_i, d) + k_1 \cdot \left(1 - b + b \cdot \frac{\text{len}}{\text{avglen}}\right)}
    \]

    \item \textbf{Semantic Search (Dense Retrieval):} \\
    Semantic similarity is estimated via the dot product between dense query and document embeddings:
    \[
    \text{Semantic score}(q, d) = q \cdot d = \sum_{i=1}^{n} q_i \cdot d_i
    \]

    \item \textbf{Reranker (Cross-Encoder):} \\
    Fine-grained semantic relevance is captured by a cross-encoder model $f_\theta$, which takes both query and document as input:
    \[
    \text{Reranker score}(q, d) = f_\theta([\text{CLS}], q, d)
    \]
\end{itemize}

To integrate the individual rankings from these components---denoted as $L$ (lexical), $S$ (semantic), and $R$ (reranker)---we adopt the Reciprocal Rank Fusion (RRF) algorithm \cite{cormack2009reciprocal}. The combined score for a document $d$ is computed as:

\[
\text{RRF}_{\text{score}_{\text{total}}} = \sum_{l \in L} \frac{1}{k + l(d)} + \sum_{s \in S} \frac{1}{k + s(d)} + \sum_{r \in R} \frac{1}{k + r(d)}
\]

where \textit{k} is a constant hyperparameter, typically set to 60 as recommended by the original paper.

In our implementation, we adopt the standard InfoNCE loss \cite{izacard2021unsupervised} as the contrastive training objective. Given a batch of queries $q_i$, their corresponding positive passages $p_i$, and a group of hard negatives corresponding to each query $\{n_{i,j}\}$, the loss is calculated as:

\[
\mathcal{L} = - \sum_{i=1}^{N} \log \frac{\exp(\mathrm{sim}(q_i, p_i)/\tau)}{\sum_{j=1}^{N} \exp(\mathrm{sim}(q_i, n_{i,j})/\tau) + \exp(\mathrm{sim}(q_i, p_i)/\tau)}
\]

where $\mathrm{sim}(\cdot)$ denotes cosine similarity and $\tau$ is a temperature parameter.

Using soft targets from a strong teacher model, our embedding model is able to internalize fine-grained semantic cues encoded in teacher output, leading to improved retrieval performance and robustness across multiple tasks.

\subsection{Context-Aware Embedding Generation via Structured Instructions}
\label{sec:method2}
Traditional embedding models typically generate representations by directly encoding input text using an encoder or a language model. While this approach is simple and computationally efficient, it lacks the flexibility to capture diverse user intents and domain-specific nuances, thereby limiting generalization capabilities. In contrast, large language models (LLMs) with decoder only, such as Mistral-7B \cite{jiang2023mistral7b}, have shown strong abilities to utilize contextual information from structured inputs. In particular, decoder-only LLMs possess strong in-context learning (ICL) capabilities, and a recent study has explored fine-tuning methods that effectively exploit the in-context learning capabilities of decoder-only LLMs \cite{li2024making}.

In our approach, we enable context-aware embedding generation by incorporating structured instructions along with few-shot examples. Each query is augmented with a task-specific instruction and a set of demonstration pairs (query-passage), designed to simulate the semantics of the target task. The input sequence is terminated with an [EOS] token, and the final embedding is extracted from the representation of the last token. This design allows the model to implicitly learn task formats, intents, and output patterns without requiring any parameter updates.

Formally, given a task instruction $t$, support examples $\{(q_i, p_i)\}_{i=1}^k$, and a target query $q$, the input is formatted as follows:

\[
\text{Examples}\{(q_i, p_i)\}_{i=1}^{k} + \text{Instruct: \{task\_definition\} + Query}
\]

The resulting embedding, extracted from the [EOS] token, reflects both the semantic content of the query and its contextual relevance to the task specification.

\subsection{Adaptive Margin-Based Mining Strategies}
\label{sec:method3}
Inspired by existing methodologies, we improve the robustness of contrastive learning and mitigate the impact of false negatives by applying adaptive margin-based mining strategies that select the negative passages of the top K according to the relevance score of the corresponding positive. This approach dynamically adjusts the threshold for negative selection, ensuring that semantically similar but nonidentical passages, often mislabeled as negatives, are excluded from training. As a result, it reduces noise and preserves the semantic contrast signal essential for effective embedding learning.

Adopting the strategy proposed by Moreira et al. \cite{moreira2024nv}, we take advantage of the ANNA IR pipeline as a teacher retrieval model to identify high-quality hard negative passages for each query. We define the maximum allowable score for negative passages as a fixed proportion of the corresponding positive score, applying a 95\% margin. This is expressed as:

\[
\text{max\_negative\_score\_threshold} = \text{positive\_score} \times \text{percentage\_of\_margin}
\]

Negative candidates with scores falling below this threshold are excluded during training. Subsequently, a random subset is sampled from the top-K ranked negatives to promote training diversity and mitigate overfitting to commonly occurring distractors. These strategies are simple yet effective and can be flexibly combined. They collectively ensure that training focuses on hard negatives that are informative but not semantically indistinguishable from the positives—thereby enhancing both retrieval accuracy and training stability.

\begin{table*}[t]
\centering
\small
\resizebox{\textwidth}{!}{
\begin{tabular}{lcccccccccc}
\toprule
\textbf{Model} & \textbf{Retrieval} & \textbf{Reranking} & \textbf{Clustering} & \textbf{PairClassification} & \textbf{Classification} & \textbf{STS} & \textbf{Summarization} & \textbf{Mean (Task)} \\
\midrule
\textit{\# of tasks} & 10 & 2 & 8 & 3 & 8 & 9 & 1 & 41 \\
\midrule
Seed1.5-Embedding               & 67.45 & 50.67 & 60.83 & 87.39 & 89.88 & 87.23 & 36.44 & 74.76 \\
gemini-embedding-001            & 64.35 & 48.59 & 59.39 & 87.70 & 90.05 & 85.29 & 38.28 & 73.30 \\
Linq-Embed-Mistral              & 60.14 & 49.44 & 54.07 & 88.44 & 83.00 & 84.69 & 37.26 & 69.80 \\
jasper\_en\_vision\_language\_v1& 56.05 & 50.00 & 60.52 & 88.14 & 90.27 & 84.37 & 37.19 & 71.41 \\
SFR-Embedding-Mistral           & 59.33 & 50.15 & 54.93 & 88.59 & 80.47 & 84.77 & 36.32 & 69.31 \\
NV-Embed-v2                     & 62.84 & 49.61 & 47.66 & 88.69 & 87.19 & 83.82 & 35.21 & 69.81 \\
text-embedding-005              & 58.77 & 48.84 & 51.91 & 87.62 & 86.03 & 85.18 & 35.05 & 69.60 \\
text-embedding-004              & 59.06 & 48.48 & 51.52 & 87.65 & 86.03 & 84.84 & 36.12 & 69.53 \\
gte-Qwen2-7B-instruct           & 58.09 & 50.47 & 58.97 & 85.90 & 88.52 & 82.69 & 35.74 & 70.72 \\
e5-mistral-7b-instruct          & 57.62 & 49.78 & 51.44 & 88.42 & 79.85 & 84.32 & 36.57 & 67.97 \\
\hline
\textbf{LGAI-Embedding-Preview(Ours)}& 66.18 & 49.13 & 59.25 & 88.67 & 89.97 & 86.69 & \textbf{38.93} & 74.12 \\
\bottomrule
\end{tabular}
}
\caption{Performance comparison of top-ranked models on the MTEB English (v2) benchmark. The models are ordered by Borda rank. The table reports average scores across 41 tasks spanning multiple categories: retrieval (10 tasks), reranking (2), clustering (8), pair classification (3), classification (8), semantic textual similarity (STS, 9), and summarization (1). Our model shows competitive results across most categories, particularly in retrieval and summarization.}
\label{tab:mainresults}
\end{table*}

\section{Experiments and Results}
\label{sec:result}
\subsection{Experiment Setup}
\label{sec:details}
\paragraph{Backbone LLM} We utilize Mistral-7B \cite{jiang2023mistral7b} as the base model and fine-tune it on a curated mixture of training datasets. Following prior works such as E5-Mistral \cite{wang2023improving} and NV-Embed-v2 \cite{lee2024nv}, we adopt Mistral-7B as the backbone of our embedding framework.

\paragraph{Fine-tuning Details} We fine-tune the Mistral-7B model over a single epoch using a contrastive loss described in \ref{sec:method1} and a learning rate of 1e-4, employing a warm-up phase followed by linear decay. To enable parameter-efficient fine-tuning, we adopt Low-Rank Adaptation (LoRA) \cite{hu2022lora}, setting the LoRA rank to 64 and the LoRA alpha to 32. In-batch negative sampling is employed for retrieval tasks, whereas it is not applied to other task types. For each instance in the retrieval datasets, 7 hard negatives are incorporated. In retrieval tasks, soft labels are generated by distilling relevance scores from the ANNA IR pipeline, which is employed as the teacher.

\paragraph{Evaluation}
The MTEB(eng, v2) benchmark was newly introduced this year as an English-only evaluation suite designed to improve computational efficiency and reduce inter-task correlation through optimized task selection \cite{enevoldsen2025mmteb}. Unlike multilingual variants, it emphasizes zero-shot evaluation by excluding tasks commonly used during fine-tuning. In this work, we target the MTEB(eng, v2) benchmark for evaluating our model’s performance. In addition, the Borda count ranking method is adopted for model comparison, following the evaluation protocol of the MTEB(eng, v2) benchmark, as it provides robust aggregation of performance across diverse tasks~\cite{colombo2022best}. For evaluation, average scores across all tasks, per-category averages, and category-weighted averages are reported. In the Borda count framework, each task is treated as a voter assigning preference rankings to models. Final rankings are computed by aggregating these votes, and in the event of ties, the tournament Borda count variant is applied to ensure ranking stability.

\subsection{MTEB Results}
\label{sec:mtebresults}
Table~\ref{tab:mainresults} presents the average performance across seven task categories—\textit{classification}, \textit{clustering}, \textit{pair classification}, \textit{re-ranking}, \textit{retrieval}, \textit{semantic textual similarity (STS)}, and \textit{summarization}—comparing our model against state-of-the-art methods reported on the MTEB leaderboard\footnote{\url{https://huggingface.co/spaces/mteb/leaderboard}}, as of June 7, 2025.

Compared to existing top-ranked models on the MTEB English (v2) benchmark, our proposed LGAI-Embedding-Preview demonstrates consistently strong performance across a broad range of task categories. Notably, it achieves high scores in both retrieval (66.18; 2nd place), sts (86.69; 2nd place), pair-classification (88.67; 2nd place) and summarization (38.93; 1st place), ranking among the top models in these categories. In particular, the model yields the highest summarization score across all evaluated systems, outperforming both larger commercial models such as \textit{gemini-embedding-001} (38.28) and \textit{Seed1.5-Embedding} (36.44). Although LGAI-Embedding-Preview reports a slightly lower mean score (74.12) compared to \textit{Seed1.5-Embedding} (74.76), it surpasses it in overall Borda ranking, reflecting stronger and more consistent performance across diverse task types. These results suggest that our model strikes a favorable balance between generalization and task-specific performance, particularly excelling in information retrieval and summarization.

\section{Conclusion}
In this report, we presented LGAI-Embedding-Preview, a unified and instruction-based framework for generating high-quality, generalized text embeddings using decoder-only LLMs. By combining in-context learning with soft supervision from a high-performance retriever-reranker pipeline and adaptive margin-based hard-negative mining, our method effectively leverages the inherent generalization capabilities of large language models without requiring full model fine-tuning or architectural changes.

Through extensive experiments on the MTEB (English v2) benchmark, our model demonstrated strong and consistent performance across a diverse set of tasks, including retrieval, classification, clustering, semantic similarity, and summarization. Notably, it achieved top-tier Borda rank, reflecting its robust generalization across both IR and non-IR settings, and ranked first in summarization despite its compact size compared to larger commercial baselines.

These findings underscore the value of structured prompting and fine-grained supervision for scalable embedding generation. Our approach highlights a promising direction for future embedding systems that prioritize efficiency, adaptability, and generalization, thereby contributing to broader real-world applications in search, recommendation, and language understanding.

We hope that this work encourages further exploration into the application of unified embedding models in real-world services, paving the way for scalable and efficient deployment of LLM-based solutions.

\bibliographystyle{unsrt}  
\bibliography{references}  






\appendix
\section*{A \quad Contributors}
All authors are listed in alphabetical order by last name.

\textbf{Core Contributors} \quad
Jooyoung Choi, Hyun Kim, Hansol Jang, Changwook Jun

\vspace{0.5em}

\textbf{Contributors} \quad
Kyunghoon Bae, Hyewon Choi, Stanley Jungkyu Choi, Honglak Lee, Chulmin Yun

\section*{B \quad Model License}

\subsection*{LGAI-Embedding-Preview Model License: Apache 2.0 and CC BY-NC 4.0}

This License Agreement (``Agreement'') is entered into between you (``Licensee'') and LG Management Development Institute Co., Ltd. (``Licensor''), governing the use of the LGAI-Embedding-Preview model (``Model''). By downloading, installing, copying, or using the Model, you agree to comply with and be bound by the terms of this Agreement. If you do not agree to all the terms, you must not download, install, copy, or use the Model. This Agreement constitutes a binding legal agreement between the Licensee and Licensor.

The LGAI-Embedding-Preview model is released under the Apache License 2.0 for software components and the Creative Commons Attribution-NonCommercial 4.0 International License (CC BY-NC 4.0) for data and model weights. For details, please refer to the full license texts at:

\begin{itemize}
  \item \url{https://www.apache.org/licenses/LICENSE-2.0}
  \item \url{https://creativecommons.org/licenses/by-nc/4.0/}
\end{itemize}

\end{document}